\documentclass[ journal]{IEEEtran}

\usepackage{ifthen}
\usepackage{amsmath,bm}
\IEEEoverridecommandlockouts       

\usepackage[utf8]{inputenc}
\usepackage{graphicx}
\usepackage{amsmath}
\usepackage{algorithm}
\usepackage[noend]{algpseudocode}
\usepackage{multirow}
\usepackage{subcaption}
\usepackage{verbatim}
\usepackage{color,soul}

\usepackage{multirow}
\usepackage[dvipsnames]{xcolor}

\DeclareMathOperator{\argmax}{argmax} 
\usepackage[switch]{lineno}
\modulolinenumbers[2]
\usepackage{cite}

\usepackage{subcaption}
\usepackage{booktabs}
\usepackage{dirtytalk}

\usepackage{lineno}

\begin{document}
\title{Boosting Exploration in Actor-Critic Algorithms by Incentivizing Plausible Novel States}
%
%
\author{Chayan Banerjee$^1$,
Zhiyong Chen$^1$, and Nasimul Noman$^2$
\thanks{$^1$School of Engineering, $^2$School of Information and Physical Sciences,
        The University of Newcastle, Callaghan, NSW 2308, Australia. 
        Z. Chen is the corresponding author. Email: {\tt\small  zhiyong.chen@newcastle.edu.au}}%
}
%
%
%
\maketitle              
\begin{abstract}
Actor-critic (AC) algorithms are a class of model-free deep reinforcement learning  algorithms, 
which have proven their efficacy in diverse domains, especially in solving continuous control problems. 
Improvement of exploration  (action entropy) and exploitation (expected return) using more efficient samples
is a critical issue in AC algorithms.
A basic strategy of a learning algorithm is to facilitate indiscriminately exploring all of the environment state space,
as well as to encourage exploring rarely visited states rather than frequently visited one.
Under this strategy, we propose a new method to boost exploration through an intrinsic reward,
based on measurement of a state's  novelty and the associated benefit of exploring the state (with regards to policy optimization), altogether called plausible novelty.
With incentivized exploration of plausible novel states, 
an AC algorithm is able to improve its sample efficiency and hence training performance. 
The new method is verified by extensive simulations of continuous control tasks of MuJoCo environments
on a variety of prominent off-policy AC algorithms.
 \end{abstract}
 

\section{Introduction}

Reinforcement learning (RL) algorithms have achieved state-of-the-art performance
in many fields including locomotion control \cite{peng2017deeploco, xie2018feedback}, autonomous driving \cite{toromanoff2020end, chen2019model}, robotics \cite{neunert2020continuous,mallick2022stochastic}, multi-agent systems and control \cite{9022871,CHEN2020109081}. 
A model-free RL algorithm bypasses the fundamentally hard problem of system modeling,
but it suffers from the issue of sample inefficiency. 
It relies solely on an agent's interactions with environment for policy learning/optimization. 
A typical strategy is to train an agent by explicitly learning a policy network aided by a concurrently learned state-value (V-value) or action-value (Q-value) network, within the actor-critic (AC) architecture \cite{konda2000actor,mnih2016asynchronous}. To boost sample efficiency, an off-policy RL algorithm maintains a so-called experience replay (ER) buffer, which stores all past samples for future reuse. In an off-policy AC algorithm, actor and/or critic networks are trained using  data uniformly sampled from an ER buffer. 


Among prominent off-policy AC algorithms, a deep deterministic policy gradient algorithm (DDPG) \cite{silver2014deterministic,lillicrap2015continuous} trains a deterministic policy network and a Q-value network (or simply called Q-network) and  it encourages exploration in action space by simply adding noise to actions. Although DDPG can achieve superior performance in certain environments, it suffers from the issue of overestimating Q-values. A twin delayed DDPG (TD3) \cite{fujimoto2018addressing} algorithm was introduced to mitigate the overestimation issue using multiple innovations including
delayed policy-update, target policy smoothing, and  a clipped double Q-value learning approach. 
As an ER buffer in off-policy algorithms needs  a large number of samples to maintain a population for meaningful policy learning,
researchers studied a variety of exploration strategies to improve sampling efficiency, such as 
action space perturbation used in DDPG and TD3, and policy parameter perturbation  in \cite{banerjee2022optimal,fortunato2017noisy,ruckstiess2010exploring}.
These strategies have their features in different environments, but they do no always perform 
well in high dimensional and sparse reward environments.
It motivates us to further study more advanced  strategies for boosting exploration in off-policy AC algorithms. 

The basic idea used in this paper is to boost exploration by adding exploration bonus, or called an intrinsic reward,  to a reward function.
Specifically, an original extrinsic reward $r_t$ is combined with an intrinsic reward $r_t^{\operatorname{intr}}$
to form an augmented reward $r_t^{\operatorname{aug}}$, denoted by
\begin{align}
r_t^{\operatorname{aug}} = r_t\oplus r_t^{\operatorname{intr}}
\end{align}
where $\oplus$ represents aggregation between the two sources of  rewards. The intrinsic reward fundamentally quantifies the novelty of a new state in an exploration process. A more novel state would receive a higher intrinsic reward and raise the probability of choosing an action to visit it.  

There are extensive researches using the idea of intrinsic reward in the literature, although not in the off-policy AC architecture.  
An intrinsic reward can be defined based on state visitation count. 
For example, the intrinsic reward in \cite{bellemare2016unifying} quantifies the count of visitations of a certain state 
using a density model and a pseudo-count generating method. 
In \cite{zhao2019curiosity}, a variational Gaussian mixture model is used to estimate densities of trajectories, 
that is, counts of visitations to a sequence of states and actions. 
In \cite{tang2017exploration}, occurrences of high dimensional states are recorded and mapped into discrete hash codes
to form intrinsic rewards. 

An intrinsic reward can also be calculated based on a prediction error method;
an exploration bonus is  awarded if there is improvement of an agent's knowledge about the environment dynamics through a predictive model. The authors of \cite{stadie2015incentivizing} 
proposed a forward dynamics model  that is trained in encoded state space using an autoencoder. 
A state's novelty and hence the intrinsic reward of visiting the state
is calculated based on the model's prediction error with respect to the state. 
The authors of \cite{pathak2019self} introduced multiple forward dynamics models and used the variance over the model outputs as an intrinsic reward. 

Another approach of obtaining  intrinsic reward is through maximization of entropy of actions/ state-action pairs / states. 
In particular, an objective function is augmented with an intrinsic reward based on a policy's entropy;
see, e.g., \cite{haarnoja2017reinforcement, ziebart2008maximum}. Zhang et. al. \cite{zhang2021exploration} uses a strategy of maximizing the Renyi entropy over state-action space for better exploration but in a reward free RL setting. In exploration phase it uses an extrinsic reward free entropy maximization objective, learns an exploratory policy and collects transition data. In the planning phase it uses the collected data and an arbitrary reward function to plan a good policy.
The basic idea of \cite{zhang2021exploration}, was further improved and  extended for a reward based RL framework in \cite{yuan2022r}. Here Renyi entropy maximization on state space is considered for improving exploration. It uses a proxy reward, which 
consists of an extrinsic reward from environment,
augmented by an action entropy maximization reward, and an additional intrinsic reward that motivates state entropy maximization.
For more extensive study on intrinsic reward or exploration bonus based techniques, the reader can refer to the surveys \cite{weng2020exploration,amin2021survey,yuan2022intrinsically} and the references therein.

Now, an interesting question is how to fit intrinsic reward based strategies into the architecture of off-policy AC algorithms. 
On one hand, assigning a reward bonus solely based on a certain novelty value indiscriminately offered for a state, as seen
in the aforementioned references,  is insufficient 
for boosting exploration in AC algorithms, because some novel states may not be worth exploring if they have a poor chance of benefiting the policy optimization process.  On the other hand,  soft actor-critic (SAC) \cite{haarnoja2018soft,haarnoja2018soft2,banerjee2022improved} is one of the most efficient action entropy maximization based algorithms for benefiting a policy optimization process.   Increase in policy's entropy results in more exploration and accelerates learning in SAC.  A physics informed intrinsic reward is proposed  in an AC algorithm in \cite{jiang2022physics},
which aims to  assist an agent to overcome the difficulty of poor training  when a reward function is sparse or misleading in short term. 
However, neither of them includes state novelty into the calculation of intrinsic rewards. 
 
The new strategy proposed in this paper focuses on exploration towards the states that have higher chances of positively impacting policy optimization, based on the measurement of a state's novelty as well as the associated benefit of exploring the state, which is called a state's plausible novelty.  
To the best of our knowledge, this is the first attempt to consider both state novelty and  benefit of exploring a state
towards policy optimization in calculating an intrinsic reward.
Another interesting feature of the new strategy is that it is an add-on/secondary artifact that can be applied to any  primary off-policy AC type algorithm, substantially improving its training performance. 
In particular, three state-of-the-art off-policy AC algorithms, SAC, DDPG, and TD3, are respectively treated as 
the primary algorithms in this paper.

The new strategy is called incentivizing plausible novel states (IPNS) throughout the paper
and it brings two major innovations. First, we propose a new state novelty scoring scheme, based on estimating a high visitation density (HVD) point from past experience. The score quantifies the Euclidean distance between a current state and an HVD point. Second, we introduce a plausible novelty (PN) score which is a combined quantification of a state's novelty score and its chance of positive contribution towards policy optimization. The chance is estimated by its V-value predicted by a concurrently trained V-network. The PN score is then normalized and weighted as an intrinsic reward bonus to be added on a primary algorithm's extrinsic reward
to form the final augmented reward, which is used in policy learning.  


%

The remaining sections of the paper are organized as follows. In Section~\ref{prelims}, we introduce preliminaries and motivations of the work. In Section~\ref{details}, we discuss  the proposed IPNS algorithm in details. Section~\ref{Experiments} contains the experiments on three MuJoco continuous control tasks and discussion about hyperparameter study, ablation study, and learning performance comparison with benchmarks. Finally we conclude this paper in Section~\ref{concl} with discussions regarding some future extension of the work.

\section{Preliminaries and motivation}\label{prelims}
We consider a  Markovian dynamical system represented by a
conditional probability density function $p(s_{t+1}|s_t,a_t)$,
where  $s_t  \in \mathcal{S}$ and $a_t \in \mathcal{A}$ 
are the state and action, respectively,  at time instant $t = 1,2,\cdots$.
 Here, $\mathcal{S}$ and $\mathcal{A}$ represent continuous state and action spaces, respectively.
Under a stochastic control policy $\pi_{\phi}(a_t|s_t)$, parameterized by $\phi$,  the distribution of the closed-loop trajectory $\tau=(s_1,a_1,s_2,a_2,\cdots,s_T, a_{T},s_{T+1})$,
over one episode $t=1,\cdots, T$, can be represented by 
\begin{align*}
   p_{\phi}(\tau) =  p(s_1)\prod_{t=1}^T 
     \pi_{\phi}(a_t|s_t) p(s_{t+1}|s_t,a_t) .
\end{align*}
Denote  $r_t = R(a_t, s_{t+1})$ as the reward generated at time $t$. 
An optimal policy is represented by the parameter
 \begin{align*}
     \phi^* = \arg\max_{\phi}\, \underbrace{{\mathbf E}_{\tau \sim p_{\phi}(\tau)}\, \Big[\sum_{t=1}^T\,\gamma ^{t} r_t \Big]}_{G_\phi},
     \end{align*}
which maximizes the objective function $G_\phi$ with a discount factor $\gamma\in (0, 1)$. 
The discounted future reward $\sum_{t=1}^T\,\gamma ^{t} r_t$ is also called a return.

The main objective of this paper is to develop a new incentive mechanism  that can be applied to boost exploration in  off-policy AC algorithms.
In this section, we use SAC as a primary off-policy AC algorithm to introduce the incentive mechanism. In Section~IV, the mechanism will be tested on 
other off-policy AC algorithms including  DDPG and TD3. 

SAC uses a maximum entropy objective, formed by augmenting the typical RL objective $G_\phi$  with the expected entropy of the policy over $p_{\phi}(\tau)$. In other words, an agent receives an intrinsic reward at each time step which is proportional to the policy's entropy at that time-step, given by   $r_t^{\operatorname{intr}} =\mathcal{H}(\pi_{\phi}(\cdot|s_{t}))$. So, the SAC's entropy-regularized RL objective to find an optimal policy can be written as
\begin{align*}
\phi^{*} = \arg\max_{\phi} \mathbf{E}_{\tau \sim p_{\phi}(\tau)} \big[ \sum_{t}^{T} \gamma ^{t}\big(r_t + \alpha  \mathcal{H}(\pi_\phi(\cdot|s_{t}))\big)\big],
\end{align*}
where the entropy regularization coefficient  $\alpha$ determines the relative importance of the entropy term against the reward. The version with a constant $\alpha$ is used in this paper. 

\begin{figure}[t]
\centering
\includegraphics[width=0.47\textwidth]{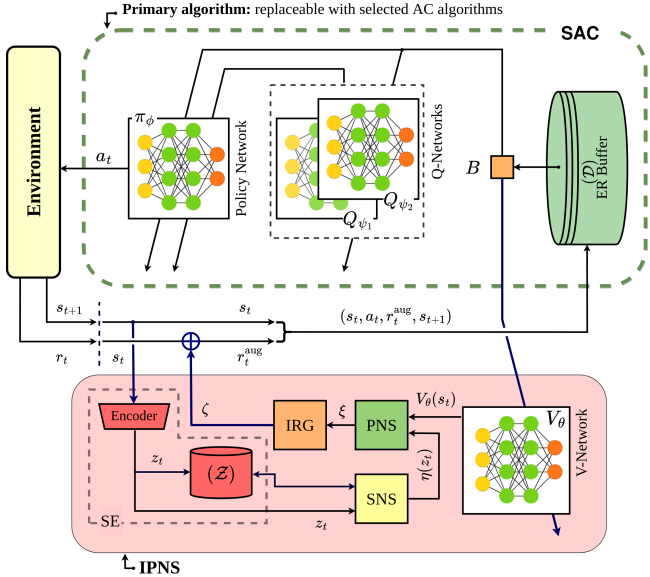}
\caption{Illustration of the proposed IPNS module paired with SAC as a primary algorithm. } 
\label{Proposed block diagram}
\end{figure}

The architecture of SAC is illustrated in Fig.~\ref{Proposed block diagram}. 
SAC learns a policy $\pi_{\phi}$ in the policy network (actor), 
which takes in the current state and generates the mean and standard deviation of an action distribution (defining a Gaussian). It concurrently learns two Q-networks  $Q_{\psi_1}, Q_{\psi_2}$ (critic) to generate Q-value, 
which assesses the expected return of a pair of state and action.
The Q-networks are learnt by regressing to the values generated by a shared pair of target networks, which are obtained by exponentially moving-averaging  the Q-network parameters over the course of training. 


As an off-policy algorithm, SAC alternates between a \say{data collection} phase and a \say{network parameter update} phase. In the data collection phase, SAC saves transition tuples $d_t^e = (s^e_t,a^e_t,r^e_t,s^e_{t+1})$
to an ER buffer  $\mathcal{D}$,  for $t = 1,\cdots,T_e, \; e= 1,\cdots,E$ ($E$ is total episodes run).
The tuples are  obtained by running the current policy in the environment. 
In the network parameter update phase, SAC uniformly samples a mini-batch ($B$) of saved transition tuples from the ER buffer $\mathcal{D}$  and updates the network parameters.
The total timesteps of the policy learning process is $N=\sum_{e=1}^E T_e$.
 
%
  
The new IPNS mechanism for boosting exploration in an off-policy AC algorithm is through defining an
intrinsic reward based on measurement of a state's novelty and the associated benefit of exploring the state,
which altogether is called plausible novelty. 
 IPNS consists of four  functional modules listed below. 
These four modules are illustrated in Fig.~\ref{Proposed block diagram} and also elaborated in the next section with 
more details.  

\begin{enumerate}
   
 \item State encoder (SE): An autoencoder is utilized to convert a state vector $s_t$ into compressed and normalized code of a reduced dimension. The encoded state vectors are stored in a new buffer $\mathcal{Z}$ that is separated from the conventional ER buffer $\mathcal{D}$.
     
\item State novelty scorer (SNS): It first estimates a point of the buffer $\mathcal{Z}$ that has HVD. 
It then calculates the novelty of a state with respect to the buffer as the Euclidean distance between the state and the HVD point.  
 
\item Plausible novelty scorer (PNS): It calculates the plausible novelty of a state by combining its state novelty and the 
benefit of exploring the state towards policy optimization. The benefit is quantified as the V-value of the state, that is, the   expected return of the state.   
    
 \item Intrinsic reward generator (IRG): It normalizes the plausible novelty of a state based on its significance in local neighborhood to generate an intrinsic reward. As a result, an augmented reward  $r_t^{\operatorname{aug}}$ by incorporating
the extrinsic reward $r_t$  with the  intrinsic reward $r_t^{\operatorname{intr}}$, will be used in policy learning algorithm.

  \end{enumerate}
    
A primary off-policy type AC algorithm, with the extrinsic reward $r_t$ replaced by the new augmented extrinsic/intrinsic reward $r_t^{\operatorname{aug}}$  following the IPNS mechanism, is enhanced to a new RL algorithm,  whose advantages in boosting its exploration capacity will be examined in various benchmark environments.  

\section{Incentivizing Plausible Novel States} \label{details}

The specific design of the four functional modules of the proposed IPNS strategy is discussed in 
the following four subsections, respectively.  
 
\subsection{State encoder (SE)} \label{AE}

The SE module includes an autoencoder and an encoded state buffer $\mathcal{Z}$.
At each timestep, the encoder performs dimensionality reduction on the original state vector $s_t \in \mathbf{R}^m$ and extracts the latent structure of the state vector in the form of a  compressed  and normalized code $z_t \in  \mathbf{R}^{m'}$ with  $m' < m$.  Dimensionality reduction is beneficial for reducing time-space complexity, simplifying distance metric calculation, 
 and hence improving learning efficiency.  More discussions about adverse effect of  high dimensional space on distance metric calculation can be found in \cite{aggarwal2001surprising}.

An autoencoder  can be defined as a deep learning algorithm that consists of a symmetrical network, with a certain hidden layer called bottleneck layer.
The left half of the network learns an encoder function $ \operatorname{enc} (\cdot)$
and generates the output $z_t = \operatorname{sig} ( \operatorname{enc} (s_t))$
through the sigmoid activation function $\operatorname{sig} (x) = \frac{1}{1+\exp(-x)}$
to further normalize  encoded state vectors.   
The right half of the network learns a decoder function $ \operatorname{dec} (\cdot)$ and 
generates  $\hat s_t =  \operatorname{dec} (z_t)$. 
 The network is trained by minimizing the loss function $L(s_t,\hat{s}_t)$ that penalizes $\hat{s}_t$ for being dissimilar to $s_t$ in the sense of mean square error.
This network training is before initiating the policy optimization process, and it uses the data collected by running a random policy in the relevant environment for certain timesteps, denoted as $N_{\rm encode}$. 
It is worth noting that these $N_{\rm encode}$ timesteps are not part of the $N$ timesteps of the policy learning process. 

During a policy learning process, the trained encoder network   encodes $s_t$ into a compressed state vector
 $z_t$, that is stored in the buffer $\mathcal{Z}$ and hence used for  state novelty score calculation as described in the next   subsection. The buffer updated as each timestep $n$ is explicitly denoted as $\mathcal{Z}_{n}$, for $n=1,\cdots, N$.
 
 

\subsection{State novelty scorer (SNS)} \label{SNS}

We first define the density of a state $z^* \in \mathcal{Z}_{n}$, denoted as $\operatorname{den}(z^*, \mathcal{Z}_{n})$. 
For this purpose, we uniformly sample $I$ mini-batches of size $L$ from the state buffer  $\mathcal{Z}_{n}$.
These mini-batches are represented by the sets 
$P^{i} = \{z_1^i, z_2^i,\cdots,z_L^i\}$ with $z_l^i \in \mathcal{Z}_{n}$, for  $l=1,\cdots,L$ and $i = 1,\cdots, I$.  
It is worth mentioning that mini-batch size is defined as $L=\operatorname{round} (\wp\%  \times n)$, where $0<\wp<100$ is a constant hyperparameter and 
the operator $\operatorname{round}(\cdot)$ returns the nearest integer. 
The density estimation formula is adapted from the density peak clustering algorithm using $K$ nearest neighbors (DPC-KNN) \cite{du2016study}. 
In particular, we define the approximate  density value of $z^*$,   estimated from the mini-batch $P^i$, as follows:
\begin{align}
\operatorname{den}_P (z^*, P^i) = e^{-\frac{1}{L}\sum_{l=1}^L \, \{\upsilon_c \| z^*-z_l^i\| \}}, \; \upsilon_c = e^{-c \| z^* - z_l^i\| }, \label{denP}
\end{align}
where $c$ is a constant and the weight $\upsilon_c$ penalizes the contribution of a datapoint in density calculation based on its distance from the $z^*$, 
measured by the Euclidean norm $\| z^* -z_l^i\|$. 
The approximate density value of $z^*$ can be repeatedly estimated from the $I$ mini-batches and forms a dataset  
$ \{ \operatorname{den}_P (z^*, P^1), \cdots, \operatorname{den}_P (z^*, P^I) \}$.
It is ready to calculate the density of $z^*$ as the average of this dataset as follows:
\begin{align}
\operatorname{den}(z^*, \mathcal{Z}_{n}) &= \frac{1}{I}\sum_{i=0}^I \operatorname{den}_P (z^*, P^i). \label{den}
\end{align}

The SNS module performs two crucial tasks. Firstly, it calculates the HVD point of the encoded state buffer  $\mathcal{Z}_{n}$, denoted as 
$\operatorname{HVD} (\mathcal{Z}_{n} )$.
For this purpose, SNS uniformly samples a set of $J$ candidate HVD datapoints from $\mathcal{Z}_{n}$, denoted as $Q = \{z^*_1,z^*_2,\cdots,z^*_J\}$ with $z^*_j \in \mathcal{Z}_{n}$, $j = 1,\cdots,J$. 
It then estimates the density $\operatorname{den}(z^*_j, \mathcal{Z}_{n})$ of each candidate datapoint and selects 
the one with the highest density as the HVD of $\mathcal{Z}_{n}$, that is, 
\begin{align}
&\operatorname{HVD} (\mathcal{Z}_{n} ) \nonumber \\
=& \left\{ \begin{array}{ll}  \argmax_{z^* \in Q} \{ \operatorname{den}(z^*, \mathcal{Z}_{n})  \},  & n= M, 2M, 3M, \cdots \\
   \operatorname{HVD} (\mathcal{Z}_{n-1} ),  & \text{otherwise}
   \end{array} \right. . \label{hvd}\end{align}
It is worth mentioning that HVD is not updated for every timestep, but for every $M$ timesteps. 
An effective $\operatorname{HVD} (\mathcal{Z}_{n} )$ can be calculated only for $n\geq M$, that is, the number of samples reaches the cut-in threshold $M$. 
With this definition, an HVD point represents the highest density zone of a state butter $\mathcal{Z}_{n}$, 
and it also  represents  clusters of states with high visitation frequency. 
    
Secondly,  SNS calculates the novelty score of  the current state $z_t$\footnote{More specifically, 
it should be denoted as $z_t^e$ that represents the current state of the $e$-th episode. 
We ignore the superscript $e$ for notation simplicity.} with respect to the buffer $\mathcal{Z}_{n}$
according to the Euclidean distance between $z_t$ and the HVD of $\mathcal{Z}_{n}$, i.e.
\begin{align}
    \eta(z_t, \mathcal{Z}_{n}) = \|z_t - \operatorname{HVD} (\mathcal{Z}_{n} ) \|.
\end{align}

Let us define the absolute density of $z^*$ in $\mathcal{Z}_{n}$  as
\begin{align*}
\operatorname{abs-den}(z^*, \mathcal{Z}_{n}) = e^{-\frac{1}{n}\sum_{z\in \mathcal{Z}_{n} } \, \{\upsilon_c \| z^*-z\| \}}, \; \upsilon_c = e^{-c \| z^* - z\| },
\end{align*}
and hence the absolute HVD of $\mathcal{Z}_{n}$ as the point with the highest absolute density, that is, 
\begin{align*}
\operatorname{abs-HVD} (\mathcal{Z}_{n} )  = \argmax_{z^* \in \mathcal{Z}_{n}} \{ \operatorname{abs-den}(z^*, \mathcal{Z}_{n})  \}. 
\end{align*}
These absolute values can be calculated by exhausting all the points in  $\mathcal{Z}_{n}$ using \eqref{denP}, 
\eqref{den}, and \eqref{hvd} with $I=1$, $L=n$, and $J=n$. However, it becomes infeasible when the size of $\mathcal{Z}_{n}$
increases.  Therefore, in our algorithms, we use only a certain amount of samples to calculate the HVD point, which is practically 
representative of high density zones.

\subsection{Plausible novelty scorer (PNS) }\label{PNS}

The PNS module is to measure the so-called plausible novelty (PN) score of 
the current state $z_t$ with respect to the buffer $\mathcal{Z}_{n}$, 
which is a combined score of the state's novelty score 
and the benefit of exploring it towards policy optimization. 
The former has been calculated as $\eta(z_t, \mathcal{Z}_{n})$ and the latter can be quantified as the V-value of the state, $V(s_t)$, which measures its expected return.    It is noted that  V-value is calculated based on the original state $s_t$ rather than the encoded state $z_t$.

 V-value is calculated by a V-network, parameterized by $\theta$ and denoted as $V_{\theta}$.
The simplest update of the network parameter vector $\theta_{n+1}  \leftarrow \theta_n$, according to the sample $(s_t, r_t,s_{t+1})$,
 can be
\begin{align*}
    V_{\theta_{n+1}}(s_t) \leftarrow V_{\theta_n}(s_t)  +\mu \delta_t,\\
    \delta_t = r_t + \gamma V_{\theta_n}(s_{t+1})-V_{\theta_n}(s_t),
\end{align*}
where $\delta_t$ is the temporal-difference (TD) error minimized using a gradient descent approach with the network learnt over time; $\mu$ is a learning rate. 
The V-network is trained by using the same samples that train the primary algorithm's policy and critic networks.
This choice is for the proposed artifacts to be easily merged with a conventional off-policy AC algorithm without 
any substantial change to its existing architecture.

As a result, the PN score of the current state ($s_t$ and $z_t$)  with respect to 
the buffer $\mathcal{Z}_{n}$, based on the V-network $V_{\theta_n}$, is defined as follows:
\begin{align}
    \xi (s_t,z_t, \mathcal{Z}_{n})= \eta(z_t, \mathcal{Z}_{n})\times V_{\theta_{n}}(s_t).
\end{align}

\subsection{Intrinsic reward generator (IRG) }\label{IRG}

The IRG module receives the PN score $\xi (s_t,z_t, \mathcal{Z}_{n})$ of the current state and normalizes it based on its significance in its local neighborhood. This normalized value is regarded as a new  
intrinsic reward, that is incorporated with the reward $r_t$ to make an augmented reward.

 For normalization of a PN score, IRG generates $K$ samples around the current state vector $z_t$, denoted as
\begin{align*}
\hat z^k_t = z_t + \rho,\; 
\rho \sim \mathcal{N}(0,0.1), \; k = 1,2,\cdots,K.
\end{align*}
Also denote the set of neighboring samples of $z_t$ as $\hat Z_t= \{\hat z_t^1,\cdots,\hat z_t^K\}$.
For each sample $\hat z^k_t$, the PN score is calculated as 
$ \xi (\operatorname{dec} (\hat z^k_t), \hat z^k_t, \mathcal{Z}_{n})$,
where the original version of  $\hat z^k_t$ is not available but it can be estimated as
$\operatorname{dec}  (\hat z^k_t)$ by the right half of the autoencoder network, i.e., the decoder. 
Then, the highest PN score of the $K$ samples is defined as 
\begin{align*}
\xi_{\max}(z_t, \mathcal{Z}_{n})= \max_{\hat z_t \in \hat Z_t} \{  \xi (\operatorname{dec} (\hat z_t), \hat z_t, \mathcal{Z}_{n})  \}  
\end{align*}
and hence the intrinsic reward   defined as
\begin{align*}
    \zeta (s_t,z_t, \mathcal{Z}_{n})  &= \frac{2}{e^{\tilde{\xi}}+e^{-\tilde{\xi}}} \\
    \tilde{\xi} &= \xi_{\max} (z_t, \mathcal{Z}_{n}) -     \xi (s_t,z_t, \mathcal{Z}_{n}).
\end{align*}
Here, $\tilde{\xi}$ is the difference between the highest PN score 
of the neighboring samples of $z_t$ and its own PN score. 
It is noted that the intrinsic reward  $\zeta (s_t,z_t, \mathcal{Z}_{n}) \in (0, 1]$. 
When the difference $\tilde{\xi} = 0$, that is, $z_t$ also receives the highest PN score of its
neighboring samples, the intrinsic reward $\zeta (s_t,z_t, \mathcal{Z}_{n})= 1$ is maximized;
otherwise,  $\zeta (s_t,z_t, \mathcal{Z}_{n}) < 1$.

For a given extrinsic reward  $r_t$ of the primary algorithm and an intrinsic reward $r_t^{\operatorname{intr}}=\zeta (s_t,z_t, \mathcal{Z}_{n})$
as calculated above, we can construct an augmented reward as follows:
\begin{align}
r_t^{\operatorname{aug}} = (1-\beta)  r_t +  \beta \zeta(s_t,z_t, \mathcal{Z}_{n}),
\end{align}
where $\beta \in [0, 1)$ is a regularization coefficient that controls the infusion of the intrinsic reward into the extrinsic reward.  
In some scenarios, an epsilon-greedy method is used to control the probability of assigning an intrinsic reward, 
that is,  \begin{align*}
r_t^{\operatorname{aug}}  = \left\{ \begin{array}{ll}  (1-\beta)  r_t +  \beta \zeta(s_t,z_t, \mathcal{Z}_{n}), & \text{probability}\; 1-\epsilon \\
r_t, & \text{probability}\; \epsilon
\end{array} \right.
\end{align*}
for $\epsilon \in [0, 1)$.  In the paper, we use $\epsilon=0$ unless a nonzero value is explicitly specified.  
It is also worth mentioning that SAC already contains an intrinsic reward proportional to the policy's entropy $\mathcal{H}(\pi_{\phi}(\cdot|s_{t}))$. When IPNS is applied on SAC, the augmented reward becomes
\begin{align*}
r_t^{\operatorname{aug}} = (1-\beta)  (r_t +  \alpha  \mathcal{H}(\pi_\phi(\cdot|s_{t})) ) + \beta \zeta(s_t,z_t, \mathcal{Z}_{n}).
\end{align*}
The three reward regularization coefficients  $\alpha, \beta, \epsilon$ are hyperparameters in a learning process
and a grid search can be adopted to determine appropriate values in different environments.

\section{MuJoco Experiments} \label{Experiments}

The discussion in this section is based on the experiments of three MuJoco \cite{6386109} continuous control tasks: InvertedDoublePendulum-v2 (InvDP), Reacher-v2 (Rch),  and Hopper-v2 (Hop).
They were implemented in OpenAI Gym \cite{brockman2016openai}
using a computer with a six-core Intel(R) Core(TM) i7-8750H CPU@2.20GHz.
The objective is to study performance improvement 
by application of the IPNS module over three prominent off-policy AC type algorithms: SAC, DDPG, and TD3.
We used the code repositories \cite{SAC2,TD3} for implementation of the respective algorithms, with 
the parameters listed in Tables~\ref{table:parameter} and \ref{table:beta_parameter}.

After a policy is trained for every certain steps (called one training unit for convenience), its performance is immediately evaluated by running the corresponding deterministic policy for five consecutive episodes. 
One training unit is $2,000$ steps for InvertedDoublePendulum-v2 and Reacher-v2, and $5,000$ steps for Hopper-v2.
There are $U= 50,100,100$ training units (the corresponding total training steps are $0.1,0.2,0.5$ million)
for InvertedDoublePendulum-v2, Reacher-v2, and Hopper-v2, respectively. 
The average return over five evaluation episodes is regarded as the episodic return $\mathcal{R}_u^\omega$ 
for the training unit $u=1,\cdots, U$. For each algorithm, this process is repeated for five times with $\omega =1,\cdots,5$.
Each repeated run is with a different random seed. 
 The same set of five random seeds was used for every pair of primary and IPNS algorithms for fair comparison. 

The autoencoder models used for different environments were trained using $N_{\rm encode}=10,000$ datapoints. These training datapoints or state vectors were obtained by running a random policy in relevant environments.
We chose an appropriate dimension of bottleneck layer such that the autoencoder's training loss $L(s_t,\hat{s}_t)$ is as low as with one significant digit $ (\sim 0.01)$.

The evolution curves of the episodic return versus the number of training units (i.e. in terms of the total number of training steps) are plotted in the figures in this section.
A solid curve indicates the mean of the five repeated runs, i.e., $\bar{\mathcal{R}}_u =  \sum_{\omega=1}^5 \mathcal{R}_u^\omega /5$
and the shaded area shows the confidence interval of the repeats representing the corresponding standard deviation $\sigma_u$. 
Each curve is smoothed using its moving average of eleven training units for clarity of understanding.

The efficiency of the IPNS strategy is discussed for InvertedDoublePendulum-v2 including HVD estimation, 
intrinsic reward calculation, ablation evaluation, and performance comparison with benchmarks in Section~\ref{sectionIDP}.
To further demonstrate the generality of efficiency, the performance of the IPNS strategy is
also compared with the benchmarks for Reacher-v2 and Hopper-v2 in  Section~\ref{sectionRchHop}.

\begin{table}[h!]
\caption{Parameters and values used in experiments.}
\centering
\begin{tabular}{ |l|l|}
\hline
Parameter & Value \\
\hline
\textbf{SAC} & \\
Learning rate & $ 3\times 10^{-4}$ \\
Optimizer & Adam \cite{kingma2014adam} \\
$\#$Hidden layers and units per layer & $ 256$\\
Experience Replay buffer size  & $10^6$\\
Soft update factor & $10^{-2}$\\
Batch size & $100$ \\
Discount factor ($\gamma$) & $0.99$ \\
Gradient step & $1$ \\
\hline 
\textbf{DDPG} & \\
Start timesteps & 1000 \\
Exploration/ action noise & 0.1 \\
$\#$Hidden layers and units per layer & 2 and 400,300 \\
\hline
\textbf{TD3} & \\
Policy noise & $0.2$ \\
Noise clip & $0.5$\\
Policy delay & $2$ \\
$\#$Hidden layers and units per layer & same as SAC \\
\hline
\textbf{IPNS} & \\
HVD update frequency ($M$) & $500$\\
$\#$Samples in IRG module ($K$) & $25$ \\
$\#$Candidate HVD datapoints($J$) & $10$ (InvDP), $5$ (Rch, Hop)\\
$\#$Mini-batches for HVD ($I$) & $100$\\
Mini-batch factor $(\wp)$ & $1$ \\
Weight multiplier ($c$) & $3$ (InvDP), $1$ (Rch, Hop)\\
$\#$Hidden layers and units per layer & same as SAC \\
(V-Network) &\\
\hline
\textbf{Autoencoder network} & \\
Encoder  &  s; D64, ELU; \\
&   D16, ELU; Sigmoid; z \\
Decoder & z; D16, ELU; D64, ELU; s  \\
Bottleneck layer dimension ($m'$) & 2 (InvDP), 5 (Rch), 3 (Hop)\\
\hline
\end{tabular}
\label{table:parameter}
\begin{flushleft}
\vspace{-0.3mm}
D stands for Dense/fully connected layers followed by number of neurons;
s and z stand for state vector and code/latent vector, respectively; and
ELU is the exponential linear unit activation function.  
\end{flushleft}
\caption{Reward regularization coefficients  $\alpha, \beta, \epsilon$.}
\centering
\begin{tabular}{ |l|l|l|l|}
\hline
Algorithm & InvDP & Rch & Hop\\
\hline
\textbf{SAC+ IPNS} & $\beta=0.1$ & $\beta=1\times 10^{-4}$ & $\beta=1 \times 10^{-3}$\\
  & $\alpha  = 0.2$  & $\alpha  = 0.2$  & $\alpha  = 0.2$  \\
\hline 
\textbf{DDPG+ IPNS} & $\beta=1\times 10^{-4}$ & $\beta=1\times 10^{-3}$ & $\beta=1\times 10^{-5}$ \\
\hline
\textbf{TD3+ IPNS} & $\beta=1\times 10^{-4}$ & $\beta= 1\times 10^{-5} $  & $\beta=1\times 10^{-4}$\\
& & $\epsilon = 0.3$ & \\
\hline
\end{tabular}
\label{table:beta_parameter}
\end{table}

\subsection{InvertedDoublePendulum-v2} \label{sectionIDP}

\subsubsection{Environment} The environment involves a cart which sits on a rail and can be moved left or right by externally applied force. A pole is fixed on the cart and a second pole is attached to the free end of the first pole. The objective here is to balance the second pole on top of the first pole by applying continuous forces on the cart. The eleven-dimensional state space includes linear position and velocity of  cart (2), angle components and angular velocities between cart, pole-1 and pole-2 (6) and constraint forces due to physical contact of robot with environment (3). The one-dimensional action represents the numerical force applied to the cart, continuous in the range of $[-1,1]$. The reward function consists of three components: (i) the agent is rewarded with +10 ($r_d$) for each timestep the seond pole is upright, (ii) a distance penalty ($p_d$) represents how far the free end of pole-2 moves, and (iii) a velocity penalty ($p_v$). Thus the overall reward function is  
\begin{align*}
    R(a_t,s_{t+1}) &= (r_d - p_d - p_v)_{t+1},  \\
    p_d &= 0.01 \times x^2 +(y-2)^2,\\
    p_v &= 0.001 \times v_1^2 + 0.005 \times v_2^2
\end{align*}
where $x$ and $y$ are the x-y coordinates of the tip of the second pole,  $v_1$ and $v_2$ are velocities of pole-1 and pole-2, respectively. 
\begin{figure}[t]
\centering
  \begin{subfigure}{.5\textwidth}
      \hspace{-0.5cm}
   \centering
  \includegraphics[width=.7\linewidth]{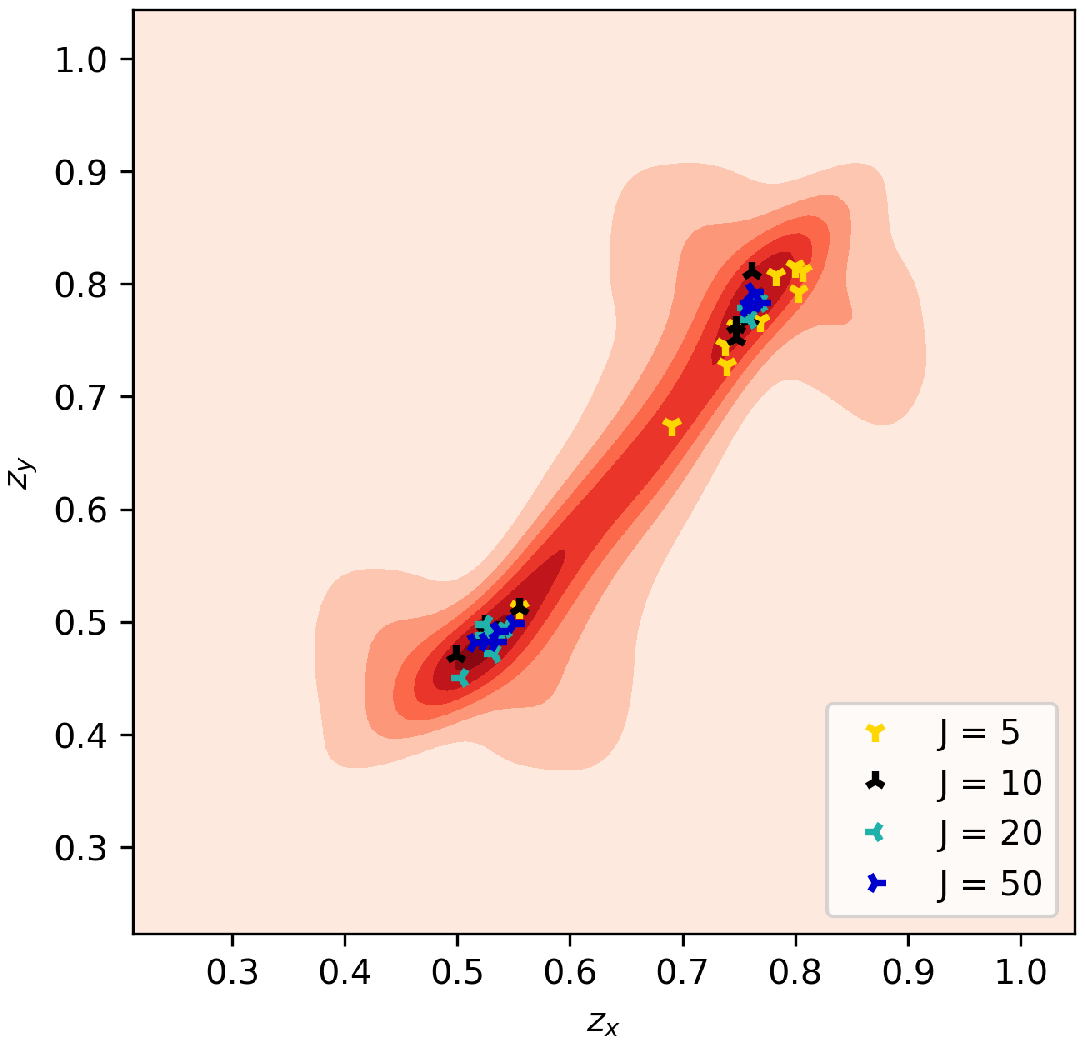}
        \vspace{-0.3cm}
   \caption{} 
   \label{param_candidate}
  \end{subfigure}
  \begin{subfigure}{.5\textwidth}
      \hspace{-0.5cm}
   \centering
  \includegraphics[width=.7\linewidth]{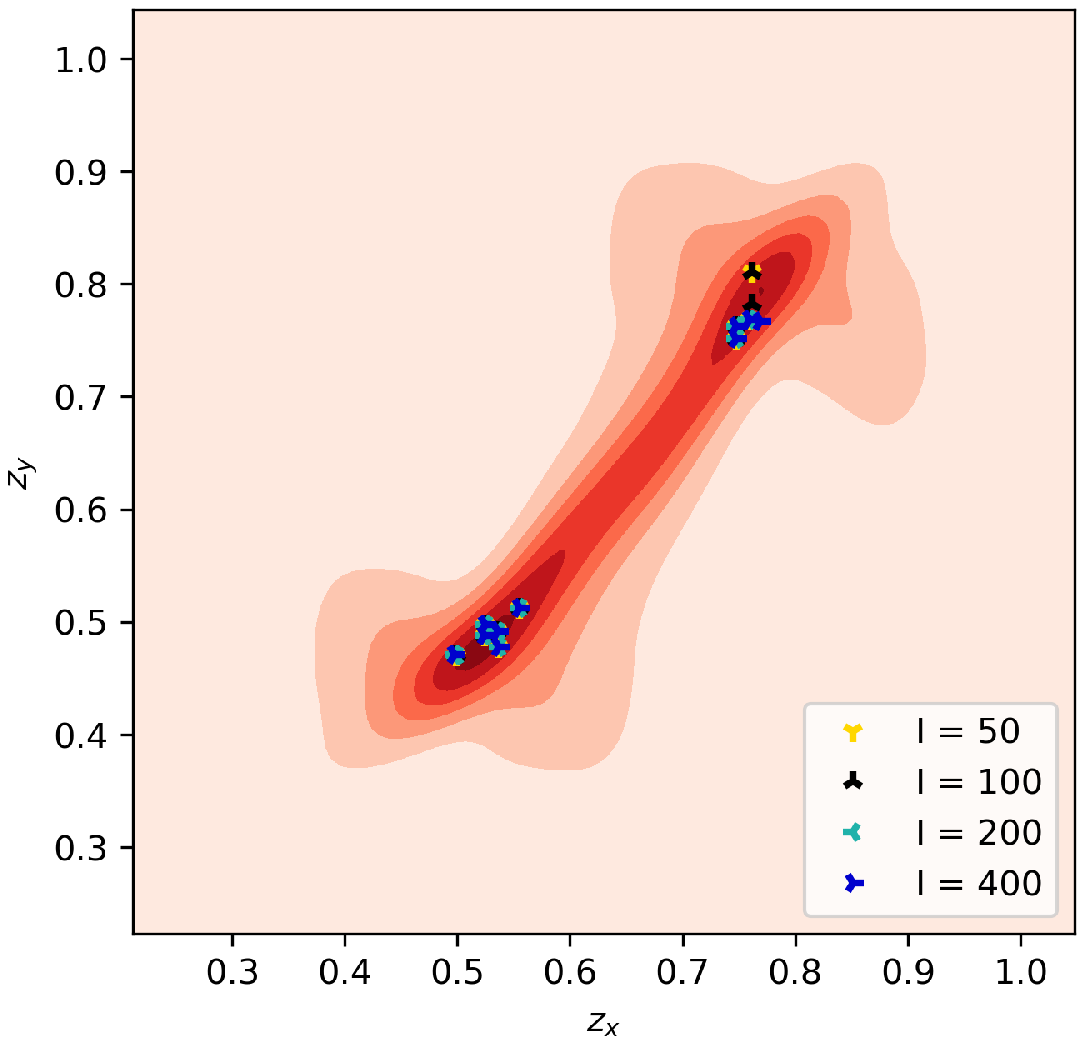} 
          \vspace{-0.3cm}
   \caption{}
   \label{param_num_minibatch}
   \end{subfigure}
  \begin{subfigure}{.5\textwidth}
      \hspace{-0.5cm}
   \centering
  \includegraphics[width=.7\linewidth]{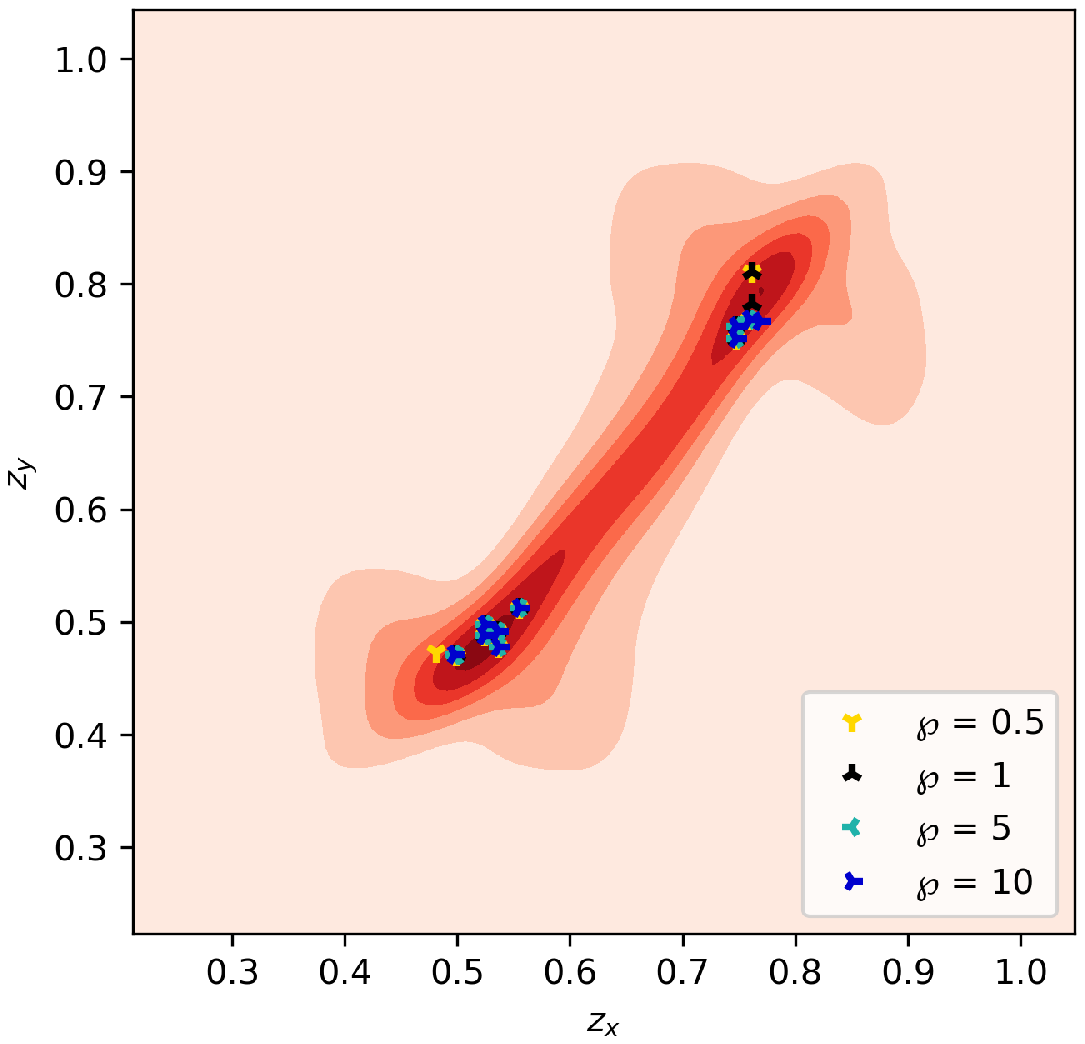} 
          \vspace{-0.3cm}
   \caption{}
   \label{param_minibatch_size}
   \end{subfigure}
\caption{HVD point estimation with different parameters: (a) $I=100$, $\wp=1$, varying $J$;
(b) $J=10$, $\wp=1$, varying $I$;
(c) $J=10$, $I=100$, varying $\wp$.}
\label{Parameter_study_HVD_estimation}
\end{figure}

\subsubsection{HVD estimation}\label{Parameter studies}

The HVD point estimation in the SNS module involves three parameters, $J$, $I$, and $\wp$.
Fig.~\ref{Parameter_study_HVD_estimation} shows the effect of changing these parameters on HVD estimation. Each  plot is based on datapoints generated by running a random policy on the InvertedDoublePendulum-v2 environment for $1\times 10^4$ timesteps.  
The underlying absolute density of every datapoint is calculated using 
 a Gaussian  kernel density estimator  \cite{rosenblatt1956remarks} and visualized in contour plots.
A comparatively higher density zone is represented in a darker color. 
The HVD values for each set of parameters are repeatedly estimated for ten times, 
which are represented by the corresponding markers in the figure. 
It is observed in all the plots that that SNS effectively estimates HVD points which 
always lie in the high density zones with various closeness to the density peaks.


For fast computational purpose, we intent to keep $J$ to a small number but not compromising with the quality of HVD estimation. It is evident from  Fig.~\ref{param_candidate} that a higher value of $J$, i.e. $J \geq 10$ makes the HVD point be well close to the two peaks. 
Also, the HVD estimations due to these values are almost similar.  Therefore, we choose $J =10$ for the current environment.
But in higher dimensional environments such as Reacher-v2 and Hopper-v2, a comparatively lower value enables IPNS to perform better. As a result, we use $J=5$ for the other two environments.  
Using the similar argument for Figs~\ref{param_num_minibatch} and \ref{param_minibatch_size}, we choose $I=100$
and $\wp=1$ for all the three environments. 

\begin{figure}[t]
\centering
  \includegraphics[width=1\linewidth]{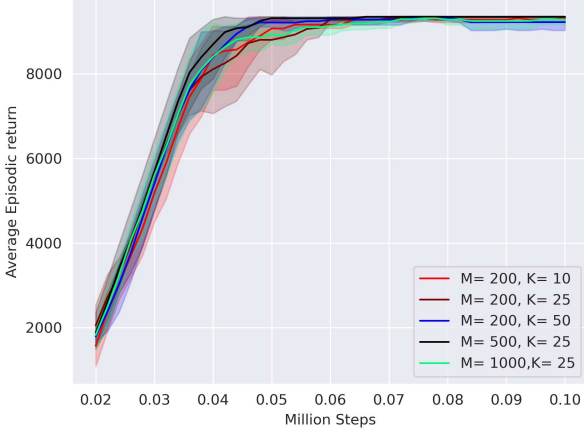}
   \caption{Training performance of SAC+IPNS with different update step size $M$ and different sample number $K$.} 
   \label{Parameter_study_MK}
\end{figure}

The calculation of HVD is updated for every $M$ timesteps. 
Training performance in the experiments with different values of $M$ is shown in Fig.~\ref{Parameter_study_MK}. 
When $M$ is relatively small, e.g., $M=200$, an HVD point is updated too frequently. 
The estimates are significantly affected by transient density distribution variations
especially when the buffer is comparatively empty, which results in 
fluctuation in estimated values over time and hence leads to slow policy learning.
For a too large $M$, e.g., $M = 1000$, an HVD point likely becomes outdated and 
poorly represents the latest data distribution, which may score a state's novelty erroneously. 
Based on these observations, we select
$M=500$ to be comparatively better for all of the environments.


\begin{figure}[t]
\centering
  \includegraphics[width=1\linewidth]{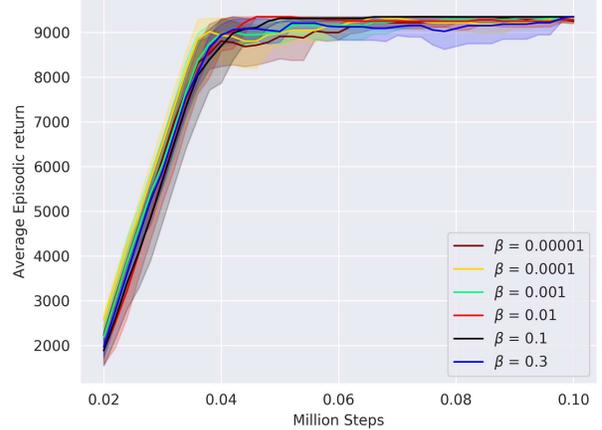}
\caption{Training performance SAC+IPNS with different intrinsic reward regularization coefficient  $\beta$. }
\label{Parameter_study_beta}
\end{figure}

\begin{figure}
    \centering
    \includegraphics[width=1\linewidth]{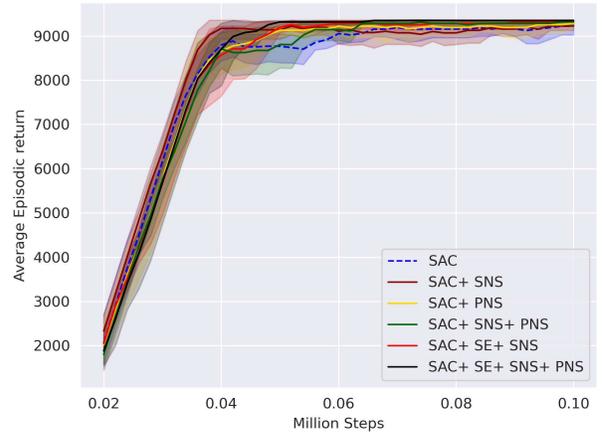}
    \caption{Ablation evaluation of individual artifacts of the IPNS algorithm.}
    \label{fig:IPNS_ablation}
\end{figure}

\subsubsection{Intrinsic reward calculation}

 IRG generates $K$ samples around the current state vector to calculate its intrinsic reward. 
We also run experiments to study the effect of $K$ on training performance. 
It is observed in Fig.~\ref{Parameter_study_MK} that a small $K=10$ leads to poor normalization of PNS score, 
resulting in learning an inferior policy; a policy with a large $K = 50$ acts myopically and shows initial rise, but 
fails to maintain its performance with time. We choose $K = 25$ that enables  good normalization and hence superior performance. 

Another important parameter for intrinsic reward assignment is the intrinsic reward regularization coefficient $\beta$. 
The influence of $\beta$ on training performance is demonstrated in Fig.~\ref{Parameter_study_beta}.
We select $\beta=0.1$ with which the best training performance is achieved by SAC+IPNS. 
The values of the three intrinsic reward regularization coefficients $\alpha, \beta, \epsilon$ in different environments
are summarized in Table~\ref{table:beta_parameter}.  For instance, as TD3+ IPNS is sensitive to $\beta$ in Reacher-v2,
a small $\beta$ and a nonzero $\epsilon$ are adopted.

\subsubsection{Ablation evaluation of IPNS}
Experiments using different modules of IPNS were conducted to study their unique contributions, as shown in Fig. \ref{fig:IPNS_ablation}. In the experiments, first the SE module is not tested separately since it is relevant only when paired with SNS for HVD estimation and state novelty calculation. 
Second, the IRG module is an essential extension to the PNS module and normalizes the PN score and hence is only relevant with PNS module and hence not tested separately in the ablation study. Third, in the case using only PNS on SAC, i.e. SAC+ PNS, the state value $V_{\theta_{t}}(s_t)$ is not weighted using the novelty score from SNS and we consider 
$\xi (s_t)= V_{\theta_{t}}(s_t)$.

We start the evaluation with simple SAC+ SNS and SAC+ PNS without SE. 
It is observed that SAC+ SNS picks up performance earlier than SAC but it fails to generalize well with time;
SAC+ PNS gives a relatively poor performance since without SNS the PN score $\xi$ is not weighted by the novelty score. 
The evaluation shows the necessity of using both modules, i.e.,  SAC+ SNS+ PNS, which 
provides better generalization over time compared to SAC+ SNS, although it  takes a long time to learn superior performance. 
Next, we examine the function of the SE module. The improvement achieved by SAC+SE+SNS over SAC+SNS shows
the efficacy of encoding states using the SE module as it gives a condensed, normalized latent representation of states
and hence leads to faster policy learning and improved generalization over time. 
Finally, adding SE to SAC+ SNS+ PNS forms a complete SAC+IPNS, which demonstrates best performance 
utilizing all the proposed artifacts.

\begin{figure}[t]
  \includegraphics[width=1\linewidth]{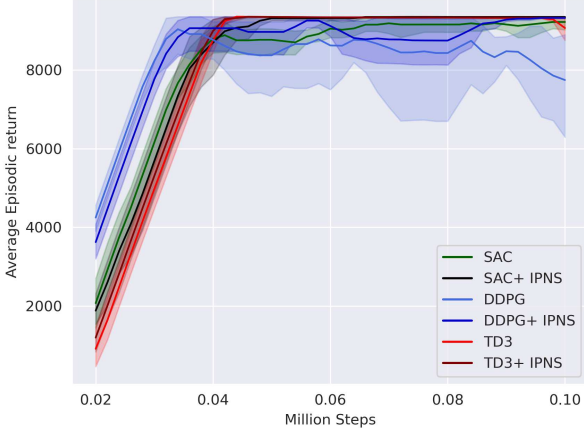}  
 \caption{Performance comparison of IPNS with other benchmarks for InvertedDoublePendulum-v2.
}
 \label{fig:compare_idp}
\end{figure}
\begin{figure}[t]
     \includegraphics[width=1\linewidth]{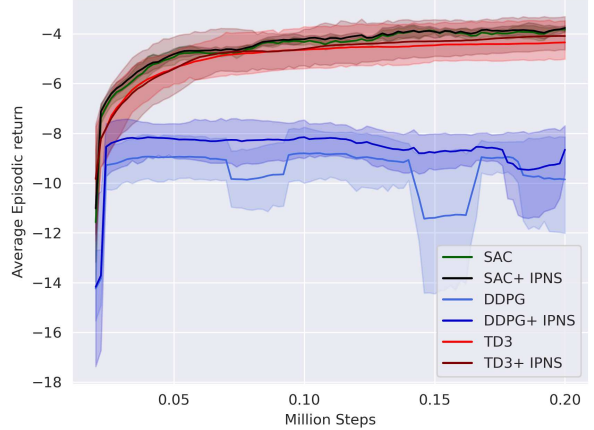} 
 \caption{Performance comparison of IPNS with other benchmarks for Reacher-v2.}
 \label{fig:compare_rch}
\end{figure}
\begin{figure}[t]
  \includegraphics[width=1\linewidth]{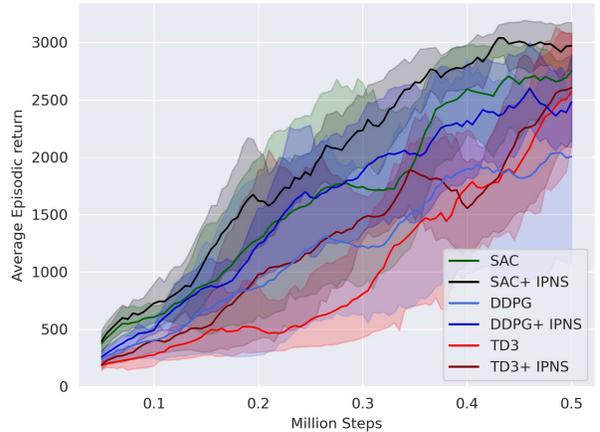} 
 \caption{Performance comparison of IPNS with other benchmarks for Hopper-v2.}
 \label{fig:compare_hop}
\end{figure}

\begin{table*}[t]
\centering
\caption{Average episodic return and standard deviation over the final 50 thousand timesteps.}
\label{table: performance max values}
\begin{tabular}{|l|l|l|l|}
\hline\hline
\textbf{Environment} & InvertedDoublePendulum-v2 &  Reacher-v2 & Hopper-v2\\ 
\hline \hline
\textbf{SAC} & $7551.895 \pm 3196.663 $   &  $ -5.108 \pm 4.841$ & $1805.805 \pm 775.995$ \\
\textbf{SAC+ IPNS} & $\mathbf{9348.266 \pm 3.661} $   &   $ \mathbf{-4.085 \pm 0.428}$ & $ \mathbf{2120.989 \pm 837.989}$\\
\textbf{DDPG} & $ 7486.696 \pm 2784.065 $  & $ -9.972 \pm 5.258$ & $1308.798 \pm 583.007$ \\
\textbf{DDPG+ IPNS} & $\mathbf{ 9072.916 \pm 534.810}$   &  $\mathbf{-8.570 \pm 0.838}$ & $\mathbf{1727.929 \pm 724.704}$ \\
\textbf{TD3} & $ 7439.094 \pm 3528.851 $  & $ -5.274 \pm 2.294 $ & $1141.182 \pm 793.818$ \\
\textbf{TD3+ IPNS} & $ \mathbf{9329.620 \pm 8.167}$   &  $\mathbf{-4.390 \pm 0.237}$ & $\mathbf{1428.622 \pm 744.101}$  \\
\hline \hline
\end{tabular}
\end{table*}

\subsubsection{Comparison with benchmarks}

The comparison is graphically exhibited in Fig. \ref{fig:compare_idp} and also quantitatively shown in Table~\ref{table: performance max values}. For comparison we calculate the average episodic return of the algorithms over their final $U_f$ units of training,  i.e., $\mathcal{R}_{f} = \sum_{u=U-U_f}^U\{ \bar{\mathcal{R}}_u \}/ U_f,$
where we use $U_f=25, 50, 10$ for InvertedDoublePendulum-v2, Reacher-v2, and Hopper-v2, respectively, all corresponding $50$ thousand training  steps.

The comparison shows that IPNS performs consistently well when paired with any of the three conventional off-policy algorithms. 
SAC+ IPNS shows superior performance in the early learning stage and retains the same performance over time, where SAC shows fluctuations and comparatively inferior performance. More specifically,  SAC+ IPNS scores $9348.266 \pm 3.661$ 
with a higher  average episodic return and a significantly lower standard deviation, compared with the SAC score $7551.895 \pm 3196.663 $. Similar improvement in scores is also observed in DDPG+ IPNS and TD3+ IPNS. 
DDPG+ IPNS improves its performance over time while DDPG fails to maintain a stable performance and deteriorates gradually.  
TD3+ IPNS shows an earlier rise compared to TD3 and retains superior performance even when TD3 dips at around $95$ thousand steps.

\subsection{Reacher-v2 and Hopper-v2} \label{sectionRchHop}

Reacher-v2 is a two-joint robot arm hinged to the centre of a square arena.   
Its two-dimensional action represents the two continuous joint torques within the range $[-1,1]$ and
its ten-dimensional state space consists of joint angles (4), coordinate of target (2), end-effector's velocity (2), and the displacement from target to end-effector  (2). The goal is to make the robot-arm's end effector reach a randomly generated target as fast as possible within one episode of $50$ steps. The reward function is $R(a_t,s_{t+1}) =  -\| d_{t+1}\|^{2} -\| a_t\|^2$  
where $d_{t+1}$, a function of the state $s_{t+1}$, is the distance between the end-effector and the target, 
and $a_t$ is the action vector.

Hopper-v2 consists of four main parts: torso (top), thigh (middle), leg (bottom), and foot. 
The goal is to hop and move in the right direction  (of x-axis), by applying torques (continuous actions in the range [-1,1]) at the three hinges that connect the four body parts. The eleven-dimensional state space consists of position (1), angles (4), and velocities (6) of the body parts.  The reward function consists of three components: (i) alive-bonus:  $+1$ for each timestep as it is alive, (ii) reward-forward: 
a reward if it hops to the right direction, (iii) reward-control: a penalty if its action value is large, represented by the following equation
\begin{align*}
    R(a_t,s_{t+1}) = 1+ \frac{x_{\rm after}-x_{\rm before}}{dt} - 1 \times 10^3 \times \|a_t \|_2^2
\end{align*}
where $x_{\rm after}$ and $x_{\rm before}$ are the x-coordinates after and before an action is taken (a function of $s_{t+1}$),  $dt$ is the time between two actions (default value: $0.008$), and $a_t$ is the action vector. One episode consists of $1,000$ timesteps but it can be terminated early primarily if (i) a hopper jumps too high, or (ii) a hopper falls, i.e., the absolute value of thigh joint is less than $0.2$ radian.

For these two environments, the IPNS strategy exhibits its improvement in learning performance 
over the benchmark AC algorithms. The results are plotted in  Fig.~\ref{fig:compare_rch}  and 
Fig.~\ref{fig:compare_hop} for  Reacher-v2 and Hopper-v2, respectively. 
They are also recorded in Table~\ref{table: performance max values}.
The improvements discussed for InvertedDoublePendulum-v2 are observed for these two environments as well. 
SAC+IPNS, DDPG+IPNS, and TD3+IPNS perform better than the corresponding primary AC algorithms in terms of both average episodic return and  standard deviation for Reacher-v2. 
We can claim that SAC+ IPNS and DDPG+ IPNS achieve 
substantial improvement in average episodic return for Hopper-v2, but with slightly more standard deviations,
compared with the corresponding primary AC algorithms.
Finally, it is observed from the plots that IPNS does not bring significant improvement when it is added to TD3
in the Hopper-v2 environment.

\section{Conclusion} \label{concl}

We have proposed a new IPNS strategy for accelerating exploration in off-policy AC  algorithms and thereby improving  sample efficiency. The key idea is to incentivize exploration towards the states of high plausible novelty scores through 
a properly designed intrinsic reward. Plausible novelty of a state consists of both state novelty and the chance of positively impacting
policy optimization by visiting the state.  
An interesting feature of IPNS is its easy implementation by integrating it with 
any  primary off-policy AC algorithm without major modification.
Three state-of-art off-policy AC algorithms have been tested as the primary algorithms
to verify the substantial improvement in learning performance by IPNS, in terms of  sample efficiency, stability,
and performance variance.  
It would be interesting to test the IPNS on a wider range of environments, especially in sparsely rewarded and multi-goal based tasks,
in future work. It would be also interesting to integrate  IPNS with other AC algorithms such as other SAC variants that dynamically adapt the exploration regularization coefficient. 

\bibliographystyle{IEEEtran}
\bibliography{refer}
\end{document}